\documentclass[11pt]{article}
\usepackage{acl}

\usepackage{times}
\usepackage{latexsym}
\usepackage{skak}
\usepackage[T1]{fontenc}
\usepackage[utf8]{inputenc}
\usepackage{microtype}
\usepackage{array}
\usepackage{booktabs}
\usepackage{graphicx}
\usepackage{scalerel}
\usepackage{amsmath}
\usepackage{amssymb}
\usepackage{multirow}
\usepackage{csquotes}
\usepackage[nameinlink]{cleveref}
\usepackage{array}
\newcolumntype{C}{@{\extracolsep{3cm}}C@{\extracolsep{0pt}}}%

\newcommand{\INF}{\ensuremath{\mathbf{\mathbb{I}}}}
\newcommand{\FOR}{\ensuremath{\mathbf{\mathbb{F}}}}
\newcommand{\INFsp}{\INF\text{ }}
\newcommand{\FORsp}{\FOR\text{ }}
\newcommand{\SUBINF}{\hspace{.1em}\scaleto{\INF}{6pt}}
\newcommand{\SUBFOR}{\hspace{.1em}\scaleto{\FOR}{6pt}}

\usepackage{todonotes}

\title{Controlling Formality in Low-Resource NMT with Domain Adaptation \\and Re-Ranking: SLT-CDT-UoS at IWSLT2022}

\author{Sebastian T. Vincent, Loïc Barrault, Carolina Scarton  \\
  Department of Computer Science, University of Sheffield \\
  Regent Court, 211 Portobello, Sheffield, S1 4DP, UK \\
  \texttt{\{stvincent1, l.barrault, c.scarton\}@shef.ac.uk}}

\begin{document}
\maketitle
\begin{abstract}
This paper describes the SLT-CDT-UoS group's submission to the first Special Task on Formality Control for Spoken Language Translation, part of the IWSLT 2022 Evaluation Campaign. Our efforts were split between two fronts: data engineering and altering the objective function for best hypothesis selection. We used language-independent methods to extract formal and informal sentence pairs from the provided corpora; using English as a pivot language, we propagated formality annotations to languages treated as zero-shot in the task; we also further improved formality controlling with a hypothesis re-ranking approach. On the test sets for English-to-German and English-to-Spanish, we achieved an average accuracy of $.935$ within the constrained setting and $.995$ within unconstrained setting. In a zero-shot setting for English-to-Russian and English-to-Italian, we scored average accuracy of $.590$ for constrained setting and $.659$ for unconstrained.
\end{abstract}

\section{Introduction}
Formality-controlled machine translation enables the system user to specify the desired formality level at input so that the produced hypothesis is expressed in a formal or informal style. Due to discrepancies between different languages in formality expression, it is often the case that the same source sentence has several plausible hypotheses, each aimed at a different audience; leaving this choice to the model may result in an inappropriate translation.

This paper describes our team's submission to the first Special Task on Formality Control in SLT at IWSLT 2022  \citep{iwslt:2022}, where the objective was to achieve control over binary expression of formality in translation (enable the translation pipeline to generate formal or informal translations depending on user input). The task evaluated translations from English (\textsc{en}) into German (\textsc{de}), Spanish (\textsc{es}), Russian (\textsc{ru}), Italian (\textsc{it}), Japanese (\textsc{ja}) and Hindi (\textsc{hi}). Among these, \textsc{en-\{ru,it\}} were considered zero-shot; for other pairs, small paired formality-annotated corpora were provided. The task ran in two settings: \textbf{constrained} (limited data and pre-trained model resources) and \textbf{unconstrained} (no limitations on either resource). Submissions within both the constrained and unconstrained track were additionally considered in two categories: full supervision and zero-shot. 

Our submission consisted of four primary systems, one for each track/subtrack combination, and we focused on the \textsc{en}-\{\textsc{de,es,ru,it}\} language directions. We were interested in leveraging the provided formality-annotated triplets $(src, tgt_{\text{formal}}, tgt_{\text{informal}})$ to extract sufficiently large annotated datasets from the permitted training corpora, without using language-specific resources or tools. We built a multilingual translation model in the given translation directions and fine-tuned it on our collected data. Our zero-shot submissions used fine-tuning data only for the non-zero-shot pairs. To boost the formality control (especially within the constrained track), we included a formality-focused hypothesis re-ranking step. Our submissions to both tracks followed the same concepts, with the unconstrained one benefitting from larger corpora, and thus more fine-tuning data.

In \Cref{constrained} we describe our submission to the constrained track, including the data extraction step (\Cref{data_collection}, \ref{dc_zero_shot}).
Our approach begins with extending this small set to cover more samples by extracting them from the allowed corpora. We use a language-independent approach of domain adaptation for this. Then, we extract samples for the zero-shot pairs (\textsc{en}-\{\textsc{ru,it}\}) based on data collected for (\textsc{en}-\{\textsc{de,es}\}). We then experiment with re-ranking the top $n$ model hypotheses with a formality-focused objective function. Within our systems, we provide the formality information as a \textit{tag} appended to the input of the model. Throughout the paper we use \FORsp to denote the \textit{formal} style and \INFsp to denote the \textit{informal} style.

All our models submitted to the \enquote{supervised} subtracks achieved an average of $+.284$ accuracy point over a baseline for all \textsc{en-\{de,es,ru,it\}} test sets, while the \enquote{zero-shot} models achieved an average improvement of $.124$ points on the \textsc{en-\{ru,it\}} test sets. Our work highlights the potential of both data adaptation and re-ranking approaches in attribute control for NMT.

\section{Constrained Track}
The MuST-C textual corpus \citep{di-gangi-etal-2019-must} with quantities listed in \Cref{tab:data_constrained} was the only data source allowed within the constrained track, alongside the IWSLT corpus of formality-annotated sentences \citep{nadejde-etal-2022-coca-mt}. MuST-C is a collection of transcribed TED talks, all translated from English. The IWSLT data itself came from two domains: telephone conversations and topical chat \citep{Gopalakrishnan2019}. The data was additionally manually annotated at phrase level for formal and informal phrases, and the organisers provided an evaluation tool \texttt{scorer.py} which, given a set of hypotheses, used these annotations to match sought formal or informal phrases, yielding an accuracy score when the number of correct matches is greater than the number of incorrect matches\footnote{\url{https://github.com/amazon-research/contrastive-controlled-mt/blob/main/IWSLT2022/scorer.py}, accessed 8 April 2022.}. This scorer skips test cases where no matches are found in the hypotheses.

\label{constrained}
\begin{table*}[h]
\centering
\scalebox{0.90}{
\begin{tabular}{@{}lcccccccc@{}}
\toprule
\multicolumn{1}{c}{\textbf{Corpus}} & \multicolumn{2}{c}{\textbf{\textsc{en-de}}} & \multicolumn{2}{c}{\textbf{\textsc{en-es}}} & \multicolumn{2}{c}{\textbf{\textsc{en-it}}} & \multicolumn{2}{c}{\textbf{\textsc{en-ru}}} \\ \midrule
MuST-C (v1.2) & \multicolumn{2}{c}{$229.7$K} & \multicolumn{2}{c}{$265.6$K} & \multicolumn{2}{c}{$253.6$K} & \multicolumn{2}{c}{$265.5$K} \\
IWSLT-22 & \multicolumn{2}{c}{$0.8$K} & \multicolumn{2}{c}{$0.8$K} & \multicolumn{2}{c}{$-$} & \multicolumn{2}{c}{$-$} \\
Formality-annotated & \FOR & \INF & \FOR & \INF & \FOR & \INF & \FOR & \INF \\
\hspace{2mm} \textsc{InferEasy} & $8.6$K & $8.6$K & $6.7$K & $6.7$K & $36.6$K & $36.6$K & $38.3$K & $38.3$K \\
\hspace{2mm} \textsc{InferFull} & $13.7$K & $9.5$K & $10.5$K & $4.5$K & $11.4$K & $13.5$K & $12.0$K & $14.1$K \\
\hspace{4mm} \textsc{+zero shot on \textsc{en-\{ru,it\}}} & $13.7$K & $9.5$K & $10.5$K & $4.5$K & $0$K & $0$K & $0$K & $0$K \\
\hspace{4mm} \textsc{+IWSLT-22} & $14.1$K & $9.9$K & $10.9$K & $4.9$K & $11.4$K & $13.5$K & $12.0$K & $14.1$K \\
\bottomrule
\end{tabular}}
\caption{Corpora containing training data used in the constrained track. Values indicate number of sentence pairs after preprocessing.}
\label{tab:data_constrained}
\end{table*}

In all our experiments we used the multilingual Transformer model architecture provided within \texttt{fairseq} \citep{ott-etal-2019-fairseq}. For our pre-training data we used the full MuST-C corpus.
We applied SentencePiece \citep{kudo-richardson-2018-sentencepiece} to build a joint vocabulary of $32$K tokens across all languages. We list the model specifications in \Cref{tab:model_spec}. Pre-training lasts $100$K iterations or $63$ epochs. We average checkpoints saved at roughly the last $10$ epochs.

\begin{table}[h]
\centering
\scalebox{0.75}{
\begin{tabular}{>{\ttfamily}l}
CUDA\_VISIBLE\_DEVICES 0,1,2,3 \\
--finetune-from-model * \\
--max-update * \\
--ddp-backend=legacy\_ddp \\
--task multilingual\_translation \\
--arch multilingual\_transformer\_iwslt\_de\_en \\
--lang-pairs en-de,en-es,en-ru,en-it \\
--encoder-langtok tgt \\
--share-encoders  \\
--share-decoder-input-output-embed  \\
--optimizer adam \\
--adam-betas '(0.9, 0.98)' \\
--lr 0.0005 \\
--lr-scheduler inverse\_sqrt \\
--warmup-updates 4000 \\
--warmup-init-lr '1e-07' \\
--label-smoothing 0.1 \\
--criterion label\_smoothed\_cross\_entropy \\
--dropout 0.3 \\
--weight-decay 0.0001 \\
--save-interval-updates * \\
-keep-interval-updates 10 \\
--no-epoch-checkpoints  \\
--max-tokens 1000 \\
--update-freq 2  \\
--fp16 
\end{tabular}}
\caption{Parameters of \texttt{fairseq-train} for pre-training and fine-tuning all models. The starred (\texttt{*}) parameters depend on the track/subtrack and can be found in the paper description or in the implementation.}
\label{tab:model_spec}
\end{table}

\subsection{Formality Controlling}
Once the model was pre-trained, we fine-tuned it on the supervised data to control the desired formality of the hypothesis with a \textit{tagging} approach \citep{Sennrich2016a}, whereby a formality-indicating tag is appended to the source input. This method has been widely used in research in various controlling tasks \citep[e.g.][]{Johnson2017, Vanmassenhove2020, Lakew2019}. 

\subsection{Automatic Extraction of Formal and Informal Data}
\label{data_collection}
Since our approach was strongly dependent on the availability of labelled data, our initial efforts focused on making the training corpus larger by extracting sentence pairs with formal and informal target sentences from the provided MuST-C corpus. 
We made the assumption that similar sentences would correspond to a similar formality level. Thus, we decided to use the data selection approach to select the most similar sentence pairs from the out-of-domain corpus (MuST-C) to both the formal and informal sides of the IWSLT corpus, which we consider our in-domain data (each side separately).

Specifically, let $G = (G_{src}, G_{tgt})$ be the out-of-domain corpus (MuST-C), and let $S_{\SUBFOR} = (S_{src}, S_{tgt, \SUBFOR})$ and $S_{\SUBINF} = (S_{src}, S_{tgt, \SUBINF})$ be the in-domain corpora (IWSLT). For simplicity, let us focus on adaptation to $S_{\SUBFOR}$. 

Our adaptation approach focuses on the target-side sentences because the IWSLT corpus is paired (for each English sentence there is a formal and informal variant in the target language). The approach builds a vocabulary of non-singleton tokens from $S_{tgt, \SUBFOR}$, then builds two language models: $LM_S$ from $S_{tgt, \SUBFOR}$ and $LM_G$ from a random sample of $10$K sentences from $G_{tgt}$; both language models use the originally extracted vocabulary. Then, we calculate the sentence-level perplexity $PP(LM_G, G_{tgt})$ and $PP(LM_S, G_{tgt})$. Finally, the sentence pairs within $G$ are ranked by
\[PP(LM_S, G_{tgt}) - PP(LM_G, G_{tgt}).\]

\noindent{Let $G_{sorted\_by\_\SUBFOR}$, $G_{sorted\_by\_\SUBINF}$ denote the resulting corpora sorted by the perplexity difference. The intuition behind this approach is that sentences which use a certain formality will naturally rank higher on the ranked list for that formality, due to similarities in the used vocabulary.}

To obtain the formal and informal corpora from the sorted data, we needed to decide on a criterion. Let $\FOR_{pos}$ and $\INF_{pos}$ be the position of a sentence pair in the formal/informal ranking, respectively. 
Our first approach was simple: let $\mathcal{C}$ denote the size of the out-of-domain corpus; we implemented an $Assign_{\theta}$ function which, for a $\theta \in [0, \mathcal{C})$, assigned a label to the sentence pair $(src, tgt)$, using the following rules: 
\[
    Assign_{\theta}
\begin{cases}
    \FOR,      & \text{if } \FOR_{pos} < \theta < \INF_{pos}; \\
    \INF,       & \text{if } \INF_{pos} < \theta < \FOR_{pos}; \\
    None,           &\text{otherwise}.
\end{cases}
\]
We condition assignment on both positional lists since common phrases such as (\textit{Yes! -- Ja!}) may rank high on both sides, but should not get included in either corpus. We determine $\theta$ empirically by selecting a value that yields the most data as a result. These values were selected dynamically for each language pair, and resulted in $\theta = 0.45\mathcal{C}$ for \textsc{en-de} and $\theta = 0.5\mathcal{C}$ for \textsc{en-es}. We refer to this approach as \textsc{InferEasy}.

We quickly observed that the selection method needed to take into account the relative ranking of a sentence pair for both formalities. To illustrate this, let $\theta=50$, the number of sentences $n=100$; a sentence pair with rankings $\FOR_{pos}=49, \INF_{pos}=51$ will get included in the formal corpus, but with $\FOR_{pos}=1, \INF_{pos}=50$ it will not, because $\INF_{pos}$ is in the top $k$ for the informal set, even though the relative difference between the two positions is large. 
To amend this, we introduced a classification by \textit{relative position difference}: for any sentence pair with positions $(\FOR_{pos}, \INF_{pos})$ we classify it as formal if $\FOR_{pos} - \INF_{pos} > \alpha$. We determine $\alpha$ empirically: using $0.05\mathcal{C}$ and $0.2\mathcal{C}$ as the lower and upper bound, respectively, for several values $\alpha$ in range we compute a language model from the resulting data and calculate average perplexity $PP(\text{LM}_{Corpus(\alpha)}, \text{IWSLT})$. We select the $\alpha$ value which minimises this perplexity. We refer to this approach as \textsc{InferFull}.

\subsection{Generalisation for Zero-Shot Language Pairs}
\label{dc_zero_shot}
For two language pairs (\textsc{en-\{ru,it\}}) no supervised training data was provided, meaning we could only use the IWSLT corpus and our inferred data from \textsc{en}-\{\textsc{de,es}\} to obtain data for these pairs. We decided to focus on comparisons on the source (\textsc{en}) side, meaning we could not use the IWSLT corpus as it was paired.
One observation we made at this point was that, contrary to intuition, the same source sentences within the MuST-C corpus had different formality expressions in the German and Spanish corpora, respectively.

Let \textsc{en-deXes} be a corpus of triplets of sentences $(src_\textsc{en}, tgt_{\textsc{de}}, tgt_{\textsc{es}})$ obtained by identifying English sentences which occur in both the \textsc{en-de} and \textsc{en-es} corpora. Since there are many such sentences in the MuST-C corpus, the \textsc{en-deXes} contains $85.72$\% of sentence pairs from the \textsc{en-de} and $74.13$\% of pairs from the \textsc{en-es} corpus. After marking the target sides of the \textsc{en-deXes} corpus for formality with \textsc{InferFull}, we quantified in how many cases both languages get the same label (formal of informal), and in how many cases they get a different label (\Cref{tab:dexes_count}). Out of all annotated triplets, only $5.8$\% triplets were annotated in both target languages; this is a significantly smaller fraction than expected. Within that group, almost 60\% triplets had matching annotations. This implies that the same English sentence can sometimes (approx. 2 out of 5 times in our case) be expressed with different formality in the target language in the same discourse situation. 

\begin{table}[h]
\centering
\begin{tabular}{@{}cccc@{}}
\toprule
\textsc{en-de} & \textsc{en-es} & Count & \% of annotated \\ \midrule
\FOR & \FOR & $845$ & $2.85\%$ \\
\INF & \INF & $233$ & $0.78\%$ \\
\FOR & \INF & $381$ & $0.95\%$ \\
\INF & \FOR & $362$ & $1.22\%$ \\
\FOR & $\emptyset$ & $10851$ & $36.54\%$ \\
\INF & $\emptyset$ & $7805$  & $26.29\%$ \\
$\emptyset$ & \FOR & $6567$ & $22.12\%$ \\
$\emptyset$ & \INF & $2749$ & $9.26\%$ \\
 \bottomrule
\end{tabular}
\caption{Context combinations for the \textsc{en-deXes} triplet extracted from the MuST-C dataset. \enquote{$\emptyset$} denotes \enquote{no context}.}
\label{tab:dexes_count}
\end{table}

Given the non-zero count of triplets with matching formalities, we make another assumption: namely that the English sentences of the triplets with matching formalities may be of \enquote{strictly formal} or \enquote{strictly informal} nature, meaning the translations of at least some of those sentences to Russian and Italian may express the same formality. To extract formal and informal sentences for the zero-shot pairs, we adapted the original method, but this time using English as a pivot to convey the formality information. As the in-domain corpus, we used the English sentences whose German and Spanish translations were both labelled as formal or both as informal, respectively (columns $1,2$ in \Cref{tab:dexes_count}). We ranked the \textsc{en-ru} and \textsc{en-it} corpora by their source sentences' similarity to that intersection (using the perplexity difference as before).


To infer the final corpora with the \textsc{InferFull} method, we used the $\alpha$ which yielded corpora of similar quantity to the ones for \textsc{en-\{de,es\}}, since we could not determine that value empirically.

\subsection{Relative Frequency Model for Reranking: \textsc{FormalityRerank}}
We observed that even when a model gets the formality wrong in its best hypothesis, the correct answer is sometimes found within the \textit{n} best hypotheses, but at a lower position. We hypothesised that by re-ranking the n-best list according to a criterion different from the beam search log probability we could push the hypothesis with the correct formality to the first position.

We performed an oracle experiment with \texttt{scorer.py} to obtain an upper bound on what can be gained by re-scoring the n-best list perfectly: we generated $k$-best hypotheses for $k \in \{1,5,10,20,30,..100\}$\footnote{We capped the search at $k=100$ due to long inference times for higher $k$ values.} and from each list of $k$ hypotheses we selected the first hypothesis (if any) which \texttt{scorer.py} deemed of correct formality. The results (\Cref{tab:oracle}) show that as we expand the list of hypotheses, among them we can find more translations of correct formality, up to a $.959$ average accuracy ($+.106$ w.r.t. the model) for $k=100$. The column \enquote{\# Cases} shows that on average in up to $21$ cases a hypothesis of the correct formality could be found with re-ranking. Finally, for any $k$, selecting the hypotheses with the correct formality (Oracle) in place of the most probable ones does (Model) not decrease translation quality, and may improve it (column \enquote{BLEU}).

\begin{table}[h]
\centering
\scalebox{0.80}{
\begin{tabular}{@{}rllcccc@{}}
\toprule
\multicolumn{1}{l}{\multirow{2}{*}{$k$}} & \multicolumn{2}{c}{Accuracy} & \multicolumn{1}{l}{\multirow{2}{*}{$\delta_{to\_best}$}} & \multicolumn{1}{l}{\multirow{2}{*}{\# Cases}} & \multicolumn{2}{c}{BLEU} \\
\multicolumn{1}{l}{} & Model & Oracle & \multicolumn{1}{l}{} & \multicolumn{1}{l}{} & Model & Oracle \\ \midrule
1 & .838 & .838 & 0.00 & 0.00 & 25.28 & 25.28 \\
5 & .858 & .892 & 1.79 & 7.00 & 24.80 & 24.80 \\
10 & .857 & .913 & 2.66 & 11.50 & 25.10 & 25.53 \\
20 & .853 & .921 & 3.46 & 13.75 & 24.74 & 25.15 \\
30 & .851 & .930 & 5.75 & 16.00 & 24.68 & 25.06 \\
40 & .853 & .936 & 7.84 & 16.75 & 24.88 & 25.24 \\
50 & .853 & .944 & 9.64 & 18.25 & 24.84 & 25.20 \\
60 & .852 & .950 & 11.78 & 19.75 & 24.71 & 25.04 \\
70 & .852 & .950 & 12.08 & 19.75 & 24.71 & 25.04 \\
80 & .852 & .952 & 12.78 & 20.25 & 24.72 & 25.04 \\
90 & .852 & .954 & 13.58 & 20.50 & 24.72 & 25.04 \\
100 & .853 & .959 & 14.66 & 21.25 & 24.72 & 25.04 \\ \bottomrule
\end{tabular}}
\caption{Results of the oracle experiment. The used model was \textit{constrained} and trained with the \textsc{InferFull} method, provided values are averaged across the development set. $\delta_{to\_best}$ describes the average distance to the first hypothesis of correct formality for cases where the most probable hypothesis is incorrect. The column \enquote{\# Cases} quantifies that phenomenon.}
\label{tab:oracle}
\end{table}


To re-rank the hypotheses we built a simple relative frequency model from the IWSLT data. 
For each term $t_i \in \mathcal{T}$ we calculated its occurrence counts $\FOR_{count}$ in the \textit{formal} set and $\INF_{count}$ in the \textit{informal} set. Let $count(t_i) = \FOR_{count}(t_i) + \INF_{count}(t_i)$.
Since we wished to focus on terms differentiating the two sets, we calculated the count difference ratio and used it as the weight $\beta$:

\[\beta(t_i) = \frac{|\FOR_{count}(t_i) - \INF_{count}(t_i)|}{\displaystyle \max_{ t_k \in \mathcal{T}} |\FOR_{count}(t_k) - \INF_{count}(t_k)|}\]
We additionally nullified probabilities for terms for which the difference of the number of occurrences in the formal and informal sets was lower than the third of total occurrences:
\[ \kappa(t_i) = \begin{cases}
    0, & \text{if } \frac{|\FOR_{count}(t_i) - \INF_{count}(t_i)|}{\FOR_{count}(t_i) + \INF_{count}(t_i)} < 0.33\footnote{Threshold was selected based on a small set of calibration terms.}; \\
    1, & \text{otherwise}
\end{cases}
\]
The probabilities could now be calculated as 
\begin{align*}
    p(\FOR|t_i) &= \frac{\FOR_{count}(t_i)}{count(t_i)} * \beta(t_i) * \kappa(t_i) \\
  p(\INF|t_i) &= \frac{\INF_{count}(t_i)}{count(t_i)} * \beta(t_i) * \kappa(t_i)
\end{align*}

For a hypothesis $Y$, a source sentence $S$ and contexts $c, \hat{c} \in \{\FOR, \INF\}, c \neq \hat{c}$, our objective function in translation thus became
\begin{align*}
    p(Y|X,c) = p(Y|X) + p(c|Y) - p(\hat{c}|Y)
\end{align*}
where
\[p(c|Y) = \sum_i p(c|y_i)\]

\begin{figure}
    \centering
    \includegraphics[width=\linewidth]{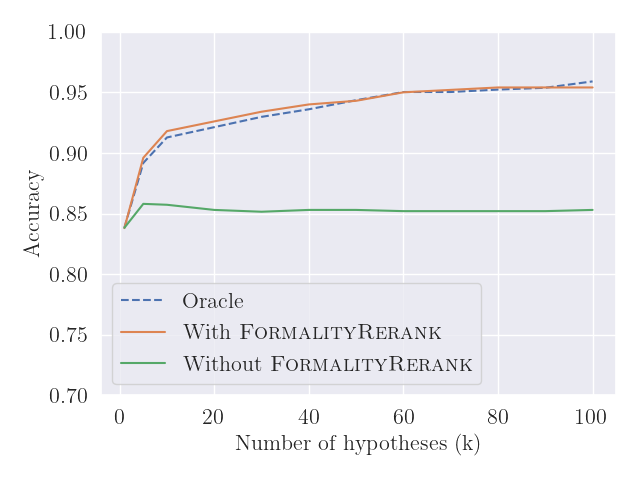}
    \caption{Validation accuracy plot showing the effect of applying \textsc{FormalityRerank} to a list of $k$ model hypotheses.}
    \label{fig:lines_oracle}
\end{figure}
\Cref{fig:lines_oracle} shows how validation accuracy increases when this method is used, and that the model is now able to match the oracle accuracy for nearly every $k$. For $k=100$ the average improvement in accuracy is $.102$. The effect of model's accuracy sometimes surpassing the oracle accuracy (e.g. for $k=30$) is a by-product of slight sample size variations: the evaluation script \texttt{scorer.py} depends on phrase matches, and a sample is only counted for evaluation if a hypothesis has at least one phrase match against the formality-annotated reference.

\subsection{Model Selection: \textsc{BestAccAveraging}}
We fine-tuned each model for $100$K iterations on the MuST-C corpus with formality tags appended to relevant sentences. We then evaluated every checkpoint (saved each epoch) with \texttt{scorer.py} on IWSLT data. 
Our initial approach to selecting a model assumed averaging the last 10 checkpoints from training. We experimented with an alternative method to finding which checkpoints to average: we first computed the accuracy on the IWSLT dataset for each checkpoint, and then selected a window of 10 consecutive checkpoints with the highest average accuracy (\textsc{BestAccAveraging}).

\subsection{Development Results}

\begin{table*}[h]
\centering
\scalebox{0.90}{
\begin{tabular}{lccccccccccc}
\toprule
 & \multicolumn{4}{c}{MuST-C (BLEU)} &  & \multicolumn{6}{c}{IWSLT (Accuracy)} \\
 & \multirow{2}{*}{\textsc{en-de}} & \multirow{2}{*}{\textsc{en-es}} & \multirow{2}{*}{\textsc{en-ru}} & \multirow{2}{*}{\textsc{en-it}} &  & \multicolumn{2}{c}{\textsc{en-de}} & \multicolumn{2}{c}{\textsc{en-es}} &  & \multirow{2}{*}{Mean} \\
 &  &  &  &  &  & \FOR & \INF & \FOR & \INF &  &  \\
 \midrule
Pre-trained & $\textbf{30.7}$ & $39.7$ & $19.5$ & $\textbf{31.3}$ &  & $.853$ & $.147$ & $.368$ & $.632$ &  & $.500$ \\
\textsc{InferEasy} & $30.1$ & $39.3$ & $19.9$ & $31.1$ &  & $.967$ & $.167$ & $.376$ & $.595$ &  & $.526$ \\
\textsc{InferFull} & $30.1$ & $\textbf{39.8}$ & $19.8$ & $31.2$ &  & $.978$ & $.637$ & $.854$ & $.963$ &  & $.858$ \\
\hspace{2mm}$+$\textsc{FormalityRerank} & $30.1$ & $\textbf{39.8}$ & $19.8$ & $31.2$ &  & $\textbf{1.000}$ & $.860$ & $\textbf{.968}$ & $\textbf{.990}$ &  & $.955$ \\
\hspace{3mm}$+$\textsc{BestAccAveraging} & $30.3$ & $39.6$ & $\textbf{20.0}$ & $31.2$ &  & $\textbf{1.000}$ & $\textbf{.899}$ & $.956$ & $\textbf{.990}$ &  & $\textbf{.961}$ \\
\bottomrule
\end{tabular}}
\caption{Results on the \textbf{development} sets for models built within the constrained track.}
\label{tab:valid_constrained}
\end{table*}

We report the validation results in \Cref{tab:valid_constrained}. The first result we observed was that in both language pairs the pre-trained model (a strong baseline) learned a \textbf{dominant} formality: formal for \textsc{en-de} ($.853$ accuracy to $.147$) and informal for \textsc{en-es} ($.632$ accuracy to $.368$). 

We observed that both methods (\textsc{InferEasy} and \textsc{InferFull}) yield consistently better accuracy for dominant formalities than non-dominant ones. Nevertheless, with \textsc{InferFull} we obtain an average $+.474$ accuracy points over the baseline for non-dominant formalities; \textsc{InferEasy} fails to learn meaningful control for non-dominant formalities. Based on these results we focused out later efforts on \textsc{InferFull} alone.

Continuing with \textsc{InferFull}, we noticed a significant improvement of up to $+.223$ accuracy points for (\textsc{en-de}, \INF) when using \textsc{FormalityRerank} on top of standard beam search ($k=100$) without impacting the translation quality. Finally, \textsc{BestAccAveraging} helped bring the average accuracy score up to $.961$ without impacting translation quality.

\subsection{Submitted Models}
Based on the validation results, we submitted two models to the constrained track: to the \textit{full supervision} subtrack, we submitted the \textsc{InferFull} model with \textsc{FormalityRerank} ($k=100$) and \textsc{BestAccAveraging} upgrades; for the \textit{zero-shot} subtrack, we fine-tuned an alternative version of the model where we skipped the \textsc{en-\{ru,it\}} fine-tuning data, effectively making inference for these zero-shot pairs\footnote{We labelled a small random sample of training data with a random formality tag so the model learned to recognise the symbol as part of the input.}. We used the same augments as in \textit{full supervision}.

\section{Unconstrained Track}
Our submission for the unconstrained track largely copies the constrained track one, but is applied to a larger training corpus.
\subsection{Data Collection and Preprocessing}

\begin{table*}[h]
\centering
\scalebox{0.90}{
\begin{tabular}{@{}ccccccccc@{}}
\toprule
\multicolumn{1}{c}{\textbf{Corpus}} & \multicolumn{2}{c}{\textbf{\textsc{en-de}}} & \multicolumn{2}{c}{\textbf{\textsc{en-es}}} & \multicolumn{2}{c}{\textbf{\textsc{en-it}}} & \multicolumn{2}{c}{\textbf{\textsc{en-ru}}} \\ \midrule
MuST-C (v1.2) & \multicolumn{2}{c}{$0.23$M} & \multicolumn{2}{c}{$0.27$M} & \multicolumn{2}{c}{$0.25$M} & \multicolumn{2}{c}{$0.27$M} \\
Paracrawl (v9) & \multicolumn{2}{c}{$278.31$M} & \multicolumn{2}{c}{$269.39$M} & \multicolumn{2}{c}{$96.98$M} & \multicolumn{2}{c}{$5.38$M} \\
NewsCommentary v16 & \multicolumn{2}{c}{$0.40$M} & \multicolumn{2}{c}{$0.38$M} & \multicolumn{2}{c}{$0.09$M} & \multicolumn{2}{c}{$0.34$M} \\
CommonCrawl & \multicolumn{2}{c}{$2.40$M} & \multicolumn{2}{c}{$1.85$M} & \multicolumn{2}{c}{$-$} & \multicolumn{2}{c}{$0.88$M} \\
WikiMatrix & \multicolumn{2}{c}{$5.47$M} & \multicolumn{2}{c}{$-$} & \multicolumn{2}{c}{$-$} & \multicolumn{2}{c}{$3.78$M} \\
WikiTitles (v3) & \multicolumn{2}{c}{$1.47$M} & \multicolumn{2}{c}{$-$} & \multicolumn{2}{c}{$-$} & \multicolumn{2}{c}{$1.19$M} \\
Europarl (v7|v10) & \multicolumn{2}{c}{$1.83$M} & \multicolumn{2}{c}{$1.97$M} & \multicolumn{2}{c}{$1.91$M} & \multicolumn{2}{c}{$-$} \\
UN (v1) & \multicolumn{2}{c}{$-$} & \multicolumn{2}{c}{$11.20$M} & \multicolumn{2}{c}{$-$} & \multicolumn{2}{c}{$-$} \\
Tilde Rapid & \multicolumn{2}{c}{$1.03$M} & \multicolumn{2}{c}{$-$} & \multicolumn{2}{c}{$-$} & \multicolumn{2}{c}{$-$} \\
Yandex & \multicolumn{2}{c}{$-$} & \multicolumn{2}{c}{$-$} & \multicolumn{2}{c}{$-$} & \multicolumn{2}{c}{$1$M} \\ 
\midrule
\textbf{Total} & & & & & & & \\
\midrule
Raw & \multicolumn{2}{c}{$291.14$M} & \multicolumn{2}{c}{$285.06$M} & \multicolumn{2}{c}{$99.23$M} & \multicolumn{2}{c}{$12.84$M} \\
Preprocessed & \multicolumn{2}{c}{$76.99$M} & \multicolumn{2}{c}{$91.29$M} & \multicolumn{2}{c}{$36.99$M} & \multicolumn{2}{c}{$3.86$M} \\
\multirow{2}{*}{Formality-annotated} & \FOR & \INF & \FOR & \INF & \FOR & \INF & \FOR & \INF \\
& $216.5$K & $187.2$K & $111.8$K & $129.7$K & $101.0$K & $172.0$K & $195.9$K & $218.4$K \\
\bottomrule
\end{tabular}}
\caption{Corpora containing training data used in the unconstrained experiments. Values indicate number of sentence pairs after preprocessing.}
\label{tab:data}
\end{table*}

We collect all datasets permitted by the organisers for our selected language pairs, including:
\begin{itemize}
    \item \textbf{MuST-C (v1.2)} \citep{di-gangi-etal-2019-must},
    \item \textbf{Paracrawl (v9)} \citep{banon-etal-2020-paracrawl},
    \item \textbf{WMT Corpora} (from the News Translation task) \citep{wmt-2021-machine}:
    \begin{itemize}
        \item \textbf{NewsCommentary (v16)} \citep{tiedemann-2012-parallel},
        \item \textbf{CommonCrawl} \citep{smith-etal-2013-dirt},
        \item \textbf{WikiMatrix} \citep{schwenk-etal-2021-wikimatrix},
        \item \textbf{WikiTitles (v3)} \citep{barrault-etal-2020-findings},
        \item \textbf{Europarl (v7, v10)} \citep{koehn2005epc},
        \item \textbf{UN (v1)} \citep{ziemski-etal-2016-united},
        \item \textbf{Tilde Rapid} \citep{rozis-skadins-2017-tilde},
        \item \textbf{Yandex}\footnote{\url{https://translate.yandex.ru/corpus?lang=en}, accessed 4 Apr 2022.}.
    \end{itemize}
    
\end{itemize}
We list data quantities as well as availability for all language pairs in \Cref{tab:data}. We preprocessed the WMT and Paracrawl corpora: for both we first ran a simple rule-based heuristic of removing sentence pairs with sentences longer than 250 tokens, and with a source-target ratio greater than 1.5; removing non-ASCII characters on the English side, pruning some problematic sentences (e.g. links).
We normalised punctuation using the script from Moses \cite{koehn-etal-2007-moses}. 
We removed cases where either sentence is empty or where the source is the same as the target. Finally, we asserted that the case (lower/upper) of the first characters must be the same between source and target and that if either sentence ends in a punctuation mark, its counterpart must end in the same one. 
As the last step, we removed identical and very similar sentence pairs.

After the initial preprocessing, we ran the \textit{BiCleaner} tool \cite{ramirez-sanchez-etal-2020-bifixer} on each corpus; the algorithm assigns a confidence score $\in [0,1]$ to each pair, measuring whether the sentences are good translations of each other, effectively removing potentially noisy sentences. We removed all sentence pairs from the corpora which scored below $0.7$ confidence. The final training data quantities are reported in \Cref{tab:data}.

\subsection{Data Labelling}
Before we applied the same method to obtain fine-tuning data for the unconstrained track, we observed that many sentence pairs in this corpus are not dialogue, and hence useless for fine-tuning. As the first step, we used the original perplexity-based re-ranking algorithm to prune the unconstrained corpus. We used the MuST-C corpus as in-domain and all the unconstrained data as out-of-domain. We truncated the unconstrained set to the top $5$M sentences most like the MuST-C data. We then applied \textsc{InferFull} with $\alpha$ threshold adapted to the data volume. The resulting data quantities can be found in the last row of \Cref{tab:data}.

\subsection{Pre-training and Fine-tuning}
We used an identical model architecture to the one from the constrained track but extended the training time: we pre-trained for $1.5$M iterations (approx. $1.5$ epochs) and fine-tuned for $0.25$M iterations (approx. $47$ epochs). For fine-tuning, we used the MuST-C corpus (to maintain high translation quality) concatenated with the inferred formality-annotated data (to learn formality control). We applied \textsc{FormalityRerank} with $k=50$, but not \textsc{BestAccAveraging} as we found that the differences in average accuracy for most checkpoints is minimal (and nears $100$); instead, we averaged the last $10$ checkpoints.

\subsection{Development Results}
\begin{table*}[h]
\centering
\scalebox{0.90}{
\begin{tabular}{lcccclcccclc}
\toprule
 & \multicolumn{4}{c}{MuST-C (BLEU)} &  & \multicolumn{6}{c}{IWSLT (Accuracy)} \\
 & \multirow{2}{*}{\textsc{en-de}} & \multirow{2}{*}{\textsc{en-es}} & \multirow{2}{*}{\textsc{en-ru}} & \multirow{2}{*}{\textsc{en-it}} &  & \multicolumn{2}{c}{\textsc{en-de}} & \multicolumn{2}{c}{\textsc{en-es}} &  & \multirow{2}{*}{Mean} \\
 &  &  &  &  &  & \FOR & \INF & \FOR & \INF &  & \\
 \midrule
Pre-trained & $28.9$ & $39.5$ & $18.5$ & $29.3$ &  & $.634$ & $.366$ & $.215$ & $.785$ &  & $.500$ \\
\textsc{InferFull} & $\textbf{32.3}$ & $\textbf{40.8}$ & $\textbf{20.4}$ & $\textbf{32.0}$ &  & $.990$ & $\textbf{1.000}$ & $.952$ & $.991$ &  & $.983$ \\
\hspace{2mm}$+$\textsc{FormalityRerank} & $\textbf{32.3}$ & $\textbf{40.8}$ & $\textbf{20.4}$ & $\textbf{32.0}$ &  & $\textbf{1.000}$ & $\textbf{1.000}$ & $\textbf{.995}$ & $\textbf{1.000}$ &  & $\textbf{.999}$ \\
\bottomrule
\end{tabular}}
\caption{Results on the \textbf{development} sets for models built within the unconstrained track.}
\label{tab:valid_unconstrained}
\end{table*}

The development results (\Cref{tab:valid_unconstrained}) surpassed those achieved in the constrained track, presumably thanks to richer corpora extracted for both formalities. \textsc{InferFull} yielded near-perfect accuracy for all sets but (\textsc{en-de}, \INF), and applying \textsc{FormalityRerank} effectively brought all scores up to a mean accuracy of $.999$. Our pre-trained model for this track achieved lower BLEU scores than for the constrained track, which is explained by the test set coming from the same domain as the constrained training data.

\subsection{Submitted model}
Similarly to the constrained track, we submit two models to the unconstrained track: to the \textit{full supervision} subtrack, we submit the \textsc{InferFull} model with \textsc{FormalityRerank} ($k=50$); for the \textit{zero-shot} subtrack, we fine-tune an alternative version of that in which we skip the \textsc{en-\{ru,it\}} fine-tuning data, effectively making inference for these pairs zero shot.
\begin{table*}[h]
\centering
\scalebox{0.90}{
\begin{tabular}{rcccclcccc}
\toprule
\multicolumn{1}{c}{\multirow{2}{*}{Model name}} & \multicolumn{4}{c}{BLEU} &  & \multicolumn{4}{c}{COMET} \\
\multicolumn{1}{c}{} & \textsc{en-de} & \textsc{en-es} & \textsc{en-ru} & \textsc{en-it} &  & \textsc{en-de} & \textsc{en-es} & \textsc{en-ru} & \textsc{en-it} \\
\midrule
\textit{constrained-supervised (1)} & $31.50$ & $36.53$ & $21.41$ & $33.28$ &  & $.4477$ & $.6076$ & $.3311$ & $.5676$ \\
\textit{constrained-zero-shot (2)} & $31.25$ & $36.65$ & $21.43$ & $33.15$ &  & $.4368$ & $.6108$ & $.3298$ & $.5525$ \\
\textit{unconstrained-supervised (3)} & $32.50$ & $36.98$ & $22.01$ & $33.56$ &  & $.4972$ & $.6349$ & $.3846$ & $.5927$ \\
\textit{unconstrained-zero-shot (4)} & $32.47$ & $36.83$ & $21.45$ & $33.12$ &  & $.4851$ & $.6209$ & $.3565$ & $.5623$ \\
\bottomrule
\end{tabular}}
\caption{Translation quality results on the \textbf{test} sets for all submitted models. Numbers in brackets indicate number of model submitted.}
\label{tab:tst_common}
\end{table*}

\begin{table*}[h!]
\centering
\scalebox{0.90}{
\begin{tabular}{@{}rcccccccc@{}}
\toprule
\multirow{2}{*}{Model name} & \multicolumn{2}{c}{\textsc{en-de}} & \multicolumn{2}{c}{\textsc{en-es}} & \multicolumn{2}{c}{\textsc{en-ru}} & \multicolumn{2}{c}{\textsc{en-it}} \\
& \FOR & \INF & \FOR & \INF & \FOR & \INF & \FOR & \INF \\
\midrule
\textit{constrained-pre-trained} & $.885$ & $.115$ & $.457$ & $.543$ & $.951$ & $.049$ & $.149$ & $.851$ \\
\textit{constrained-supervised (1)} & $1.000$ & $.886$ & $.874$ & $.980$ & $.981$ & $.234$ & $.349$ & $.961$ \\
\textit{constrained-zero-shot (2)} & $-$ & $-$ & $-$ & $-$ & $.981$ & $.154$ & $.294$ & $.929$ \\
\addlinespace[0.5em]
\textit{unconstrained-pre-trained} & $.745$ & $.255$ & $.323$ & $.677$ & $.964$ & $.036$ & $.052$ & $.948$ \\
\textit{unconstrained-supervised (3)} & $1.000$ & $1.000$ & $.981$ & $1.000$ & $.992$ & $.136$ & $.188$ & $.980$ \\
\textit{unconstrained-zero-shot (4)} & $-$ & $-$ & $-$ & $-$ & $.995$ & $.142$ & $.512$ & $.986$ \\
\bottomrule
\end{tabular}}
\caption{Accuracy results on the \textbf{test} data as measured by \texttt{scorer.py}.}
\label{tab:tst_acc}
\end{table*}

\section{Final Results}
We report the final evaluation results in \Cref{tab:tst_common} (translation quality) and \Cref{tab:tst_acc} (formality control). In the latter we also provide the performance of our baseline (pre-trained) model for reference.

Within the constrained track, we achieved near-ideal accuracy for the dominant formality for each language pair (between $.961$ and $1.000$) with the supervised model. 
Scores for non-dominant formalities are weaker but still impressive for \textsc{en-\{de,es\}} with an average of $.880$. Our best model for \textsc{en-\{ru,it\}} improved by $.193$ accuracy points over the baseline. 
The models submitted to the unconstrained track again achieved an impressive average accuracy of $.992$ for dominant formality; additionally, performance for non-dominant formality in \textsc{en-\{de,es\}} improved significantly w.r.t. the constrained model, also averaging $.992$. This means that with enough training data our methods were capable of matching the performance on a minority class w.r.t. a majority class. 

Finally, contrary to the constrained track, the \textit{unconstrained-zero-shot} model achieved the best accuracy for zero-shot pairs, to an average of $.659$.

\section{Conclusions}
Overall results suggest that it is easy for a pre-trained translation model to learn controlled expression of the dominant type within a dichotomous phenomenon while learning to render the less-expressed type is significantly harder, especially in a low-resource scenario. Our methods applied to the supervised language pairs (English-to-German, English-to-Spanish) worked near unfailingly, but using English as a pivot language to propagate formality information did not help achieve similar results for the zero-shot pairs. 

We suspect that the significant accuracy gains from \textsc{FormalityReranking} may have been partially due to formality in the studied language pairs itself being expressed primarily via certain token words such as the honorific \textit{Sie} in German creating a \textit{pivot} effect \citep{fu-etal-2019-rethinking}. As such, it may be of interest for future research to study such methods applied to more complex phenomena, such as grammatical expression of gender.

Finally, results for the \textsc{en-\{ru,it\}} language pairs may not have been as good as expected because we used the inferred data from the constrained track to build the relative frequency model, but the inferred data turned out to be not as high quality as we expected. Future work may investigate a robust solution to this problem of propagating formality via a source (pivot) language to extract training data for other language pairs.

Code used for our implementation can be accessed at \url{https://github.com/st-vincent1/iwslt_formality_slt_cdt_uos/}.

\section*{Acknowledgements}
This work was supported by the Centre for Doctoral Training in Speech and Language Technologies (SLT) and their Applications funded by UK Research and Innovation [grant number EP/S023062/1].

\bibliography{anthology,references}
\end{document}